\documentclass[11pt]{article}

\usepackage{arxiv}
\usepackage[utf8]{inputenc}
\usepackage[T1]{fontenc}
\usepackage{hyperref}
\usepackage{url}
\usepackage{booktabs}
\usepackage{amsfonts}
\usepackage{nicefrac}
\usepackage{microtype}
\usepackage{lipsum}
\usepackage{graphicx}

\title{GLU Attention Improve Transformer}

\author{
Wang Zehao \\
Nanjing University \\
\texttt{wangzehao\_ai@163.com}
}

\begin{document}
\maketitle

\begin{abstract}

Gated Linear Units (GLU) have shown great potential in enhancing neural network performance. In this paper, I introduce a novel attention mechanism called GLU Attention, which introduces nonlinearity into the values of Attention. My experiments demonstrate that GLU Attention improves both model performance and convergence speed across text and vision modalities with zero additional parameters and negligible computational costs. GLU Attention is lightweight and can seamlessly integrate with other technologies, such as Flash Attention, Rotary Position Embedding (RoPE), and various Multi-Head Attention (MHA) variants such as Grouped-Query Attention (GQA). This project is open-sourced at github\cite{wangzehao2025gluattentiongithub}.
\end{abstract}

\section{Introduction}

Transformer\cite{vaswani2023attentionneed} models have become the foundation of modern artificial intelligence. Transformer is a sequence-to-sequence model that uses Attention layer to capture relationships between tokens and Feed Forward Network (FFN) layer to perform transformations on each token. GLU FFN\cite{shazeer2020gluvariantsimprovetransformer} outperforms the original FFN and has been adopted in popular open source Large Language Model (LLM) Llama 3\cite{grattafiori2024llama3herdmodels}. My study shows GLU Attention outperforms original MHA. In MHA there is a softmax function introduce nonlinearity for querys and keys, but the values are projected by linear transformations. My study explores the integration of GLU nonlinearity into the values of MHA. Experiments show that adding GLU to MHA values can significantly enhance both model performance and training efficiency, making GLU Attention a simple yet effective improvement to the Transformer architecture.

\section{Backgrounds}

\subsection{Gated Linear Units}

GLU were first introduced to improve performance by introducing nonlinearity and has been successfully applied in various architectures, including convolutional neural networks (CNN)\cite{dauphin2017languagemodelinggatedconvolutional} and transformer FFN layer\cite{shazeer2020gluvariantsimprovetransformer}.

GLU contains two inputs: a gate $g$ and a gated input $x$, along with one Rectified Linear Unit\cite{fukushima1975relu} (ReLU) $ReLU(x)=max(0,x)$ or a ReLU like activation function such as Sigmoid-weighted Linear Unit (SiLU) $SiLU(x)=x*sigmoid(x)$. In this paper I use SiLU\cite{elfwing2017sigmoidweightedlinearunitsneural} as the activation function, GLU is defined as:

\begin{equation}
GLU(x,g) = x*SiLU(g)
\end{equation}
Or just split the last dimension of input $x$ into two parts, $x_1$ and $x_2$, and apply SiLU to $x_2$:
\begin{equation}
x_1, x_2 = split(x,dim=-1)
\end{equation}
\begin{equation}
GLU(x) = x_1 * SiLU(x_2)
\label{eq:glu}
\end{equation}

\subsection{Multi-Head Attention}

MHA is a key component of the Transformer architecture, enabling the model to focus on different parts of the input sequence simultaneously. The MHA layer has three inputs: queries $Q$, keys $K$, values $V$, and one output $O$. MHA applies three linear transformations $W_Q$, $W_K$, and $W_V$ to project the inputs into different subspaces for each attention head. A final linear transformation $W_O$ is used to project the output back to the original space. The MHA layer can be expressed as:

\begin{equation}
Q' = W_Q(Q)
\end{equation}
\begin{equation}
K' = W_K(K)
\end{equation}
\begin{equation}
V' = W_V(V)
\label{eq:mha_v}
\end{equation}
\begin{equation}
O' = MHA(Q', K', V')
\end{equation}
\begin{equation}
O = W_O(O')
\end{equation}

In Multi-Head Self-Attention, the same input $X$ is used for queries, keys, and values. $Q = K = V = X$

\begin{figure}[htbp]
    \centering
    
    \begin{minipage}{0.65\textwidth}
    \includegraphics[width=\textwidth]{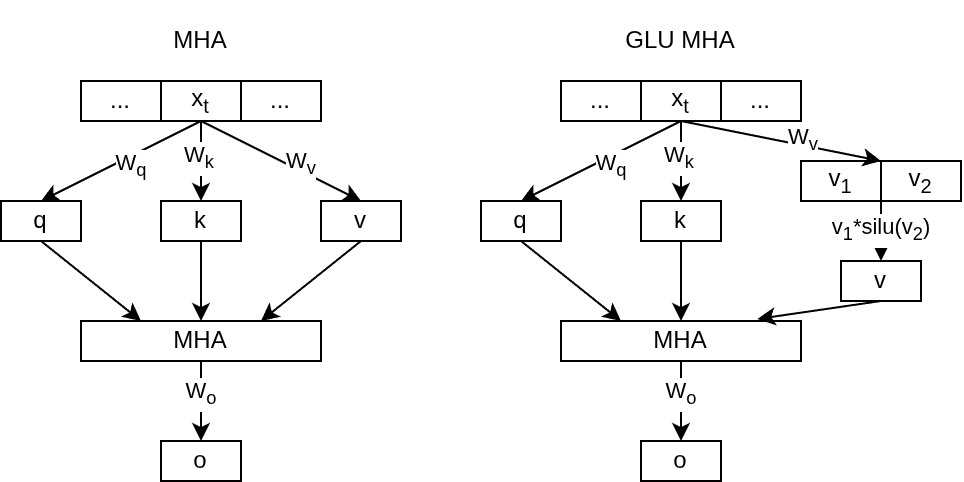}
    \caption{MHA and GLU MHA algorithm}
    \end{minipage}
\end{figure}

\section{Method}
GLU Attention uses the projected value $V'$ as the intput of Equation (\ref{eq:glu}). The GLU Attention can be expressed as:

\begin{equation}
V' = GLU(W_V(V))
\label{eq:glu_v}
\end{equation}

By replacing Equation (\ref{eq:mha_v}) in MHA with Equation (\ref{eq:glu_v}), while keeping other components unchanged, we obtain GLU Multi-Head Attention.

\section{Experiments}

\subsection{Models and Hyperparameters}

I conducted experiments using two Transformer models: a baseline model with standard Multi-Head Attention (MHA), and a GLU Attention model with GLU Multi-Head Attention (GLU MHA). Both models consist of one embedding layer, one positional embedding layer, six transformer layers, and one classification layer. Each transformer layer contains a self-attention mechanism and a GLU feed-forward network (GLU FFN), with a model dimension of 384 and 8 attention heads.

To ensure a fair comparison, the GLU MHA projection layers are designed to match the number of parameters and computational costs of classic MHA. In the GLU Attention model, the value projection layer has a shape of 384→512, and the output projection layer has a shape of 256→384, while other projection layers maintain a shape of 384→384.

The GLU FFN consists of two linear layers: the first layer has a shape of 384→2048, and the second layer has a shape of 1024→384.

\subsection{Cifar-10}

\begin{figure}[htbp]
    \centering
    \begin{minipage}{0.49\textwidth}
        \centering
        \includegraphics[width=\textwidth]{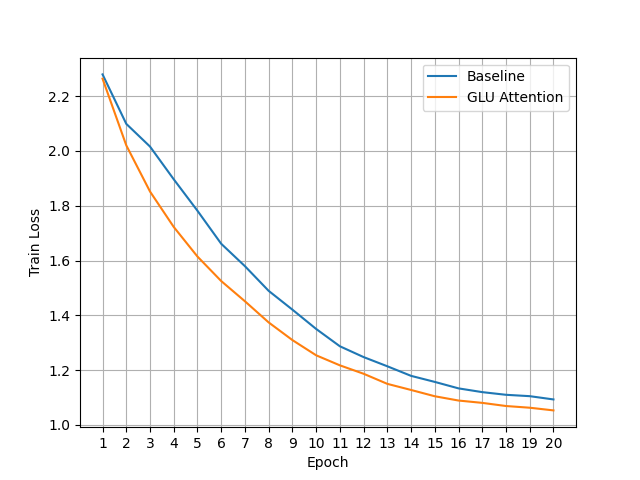}
        \caption{Cifar-10 training loss of each epoch. The lower the better.}
        \label{fig:cifar10_train_loss}
    \end{minipage}
    \hfill
    \begin{minipage}{0.49\textwidth}
        \centering
        \includegraphics[width=\textwidth]{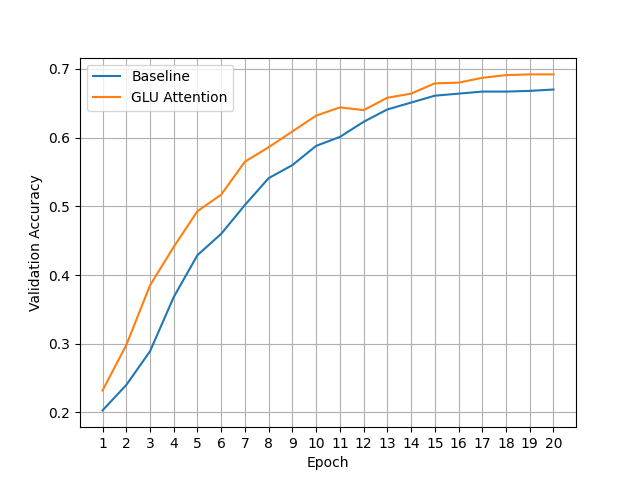}
        \caption{Cifar-10 validation accuracy of each epoch. The higher the better.}
        \label{fig:cifar10_val_acc}
    \end{minipage}
\end{figure}

I trained these models from scratch on the Cifar-10 dataset, a widely used benchmark for image classification. The training dataset consists of 60,000 32x32 color images across 10 classes, while the validation set consists of 10,000 images. I followed the standard ViT\cite{dosovitskiy2021imageworth16x16words} procedure, dividing each 32x32x3 image into 64 patches of size 4x4x3. Training was conducted for 20 epochs with a batch size of 384. I used the AdamW optimizer with a learning rate of 1e-4 and a cosine annealing scheduler. The results are shown in Figure \ref{fig:cifar10_train_loss} and Figure \ref{fig:cifar10_val_acc}. GLU Attention consistently outperformed the baseline model.

\begin{figure}[htbp]
    \centering
    \begin{minipage}{0.49\textwidth}
        \centering
        \includegraphics[width=\textwidth]{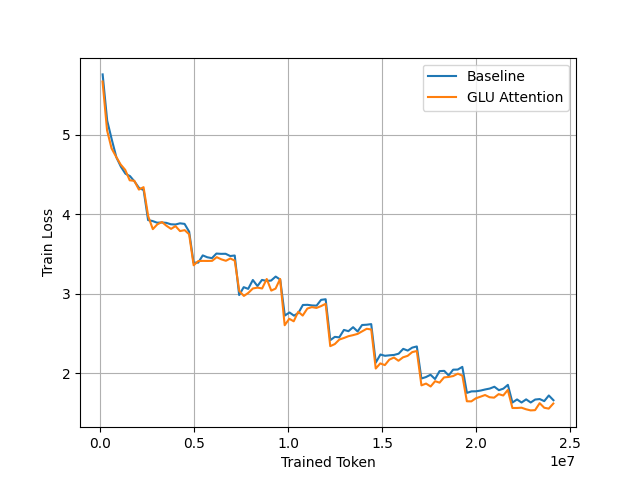}
        \caption{wikitext2 training loss for 10 epochs. The lower the better.}
        \label{fig:wikitext2_train_loss}
    \end{minipage}
    \begin{minipage}{0.49\textwidth}
        \centering
        \includegraphics[width=\textwidth]{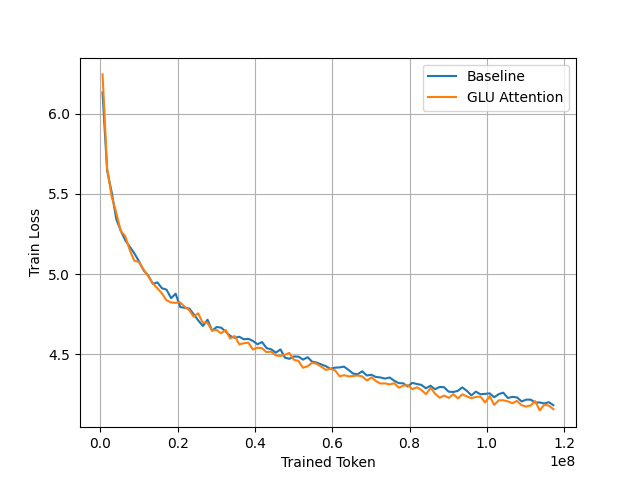}
        \caption{wikitext103 training loss for 1 epoch. The lower the better.}
        \label{fig:wikitext103_train_loss}
    \end{minipage}
    \hfill
\end{figure}

\subsection{WikiText-2}

I also trained these models from scratch on the WikiText-2 dataset which has 36,718 rows of token for language model pre-training, which is to predict the next token. I used the GPT-2 tokenizer to tokenize the text and applied the same training settings as in Cifar-10, except that the batch size was set to 1 and a causal mask was used to prevent the model from seeing future tokens. Training was conducted for 10 epochs. The results are shown in Figure \ref{fig:wikitext2_train_loss}. GLU Attention consistently outperformed the baseline model.

\subsection{WikiText-103}

Then I trained these models from scratch on the WikiText-103 dataset which has 1,801,350 rows of token using learning rate 1e-5 for 1 epoch. The results are shown in Figure \ref{fig:wikitext103_train_loss}. GLU Attention consistently outperformed the baseline model.

\section{Conclusion}

GLU Attention offers a straightforward yet impactful improvement to the Transformer architecture. By introducing nonlinearity into the values of MHA, it enhances model performance and convergence speed.

GLU Attention can be seamlessly integrated with other technologies, such as Flash Attention\cite{dao2022flashattentionfastmemoryefficientexact}, RoPE\cite{su2023roformerenhancedtransformerrotary}, and various MHA variants like MQA and GQA\cite{ainslie2023gqatraininggeneralizedmultiquery}, by simply adding a GLU function (Equation \ref{eq:glu}) after the value projection function and adjusting some parameters to accommodate the GLU function's property that output dimension is half of the input dimension.

\section{Future Work}
I highly recommend every researcher to test GLU Attention in your Transformers, as it is easy to adopt and provides a nearly cost-free performance boost. Future work could explore its application in different MHA variants with different FFN variants, on more datasets and tasks, as well as its scalability to larger models and datasets.

\bibliographystyle{unsrt}
\bibliography{references}

\end{document}